%% file: main.tex
\documentclass[pmlr]{jmlr}

\RequirePackage{graphicx}
 \usepackage{booktabs}
 \usepackage{amsmath}
 \usepackage{amssymb}
 \usepackage{bbm}
\usepackage{longtable}
\usepackage{subcaption}
\makeatletter
\def\set@curr@file#1{\def\@curr@file{#1}} 

\theorembodyfont{\upshape}
\theoremheaderfont{\scshape}
\theorempostheader{:}
\theoremsep{\newline}

\title[Ganatra \& Goel]{PRECISe : Prototype-Reservation for Explainable Classification under Imbalanced and Scarce-Data Settings}

\author{\Name{Vaibhav Ganatra}
           \Email{T-VAGANATRA@MICROSOFT.COM}\\ 
       \addr Microsoft Research\\
       India 
       \AND
       \Name{Drishti Goel}
       \Email{T-DRGOEL@MICROSOFT.COM}\\ 
       \addr Microsoft\\
       India} 

\begin{document}

\maketitle

\begin{abstract}
  Deep learning models used for medical image classification tasks are often constrained by the limited amount of training data along with severe class imbalance. Despite these problems, models should be explainable to enable human trust in the models' decisions to ensure wider adoption in high risk situations. In this paper, we propose PRECISe, an explainable-by-design model meticulously constructed to concurrently address all three challenges. Evaluation on 2 imbalanced medical image datasets reveals that PRECISe outperforms the current state-of-the-art methods on data efficient generalization to minority classes, achieving an accuracy of $\sim$87\% in detecting pneumonia in chest x-rays upon training on $<$ 60 images only. Additionally, a case study is presented to highlight the model's ability to produce easily interpretable predictions, reinforcing its practical utility and reliability for medical imaging tasks.
\end{abstract}

\section{Introduction}
\input{sections/introduction}

\section{Related Work}
\input{sections/relatedWork}

\section{Methods}
\input{sections/methodology}

\section{Experiments and Results}
\input{sections/experiments}

\input{sections/results}

\input{sections/discussions}

\section{Conclusion}
In this paper, we proposed PRECISe, an explainable-by-design model which performs well with limited and imbalanced data. We extensively evaluate the model, in various aspects, such as overall-performance, performance on minority classes, data-efficiency as well as the explainations provided by the model. We hope that the proposed method be a first step in the development of holistic-models with a deployment-first mindset, i.e. models which simultaneously tackle multiple problems associated with automated medical image analysis. Future directions of the proposed method include a more thorough examination of the explanation provided by the model, and examining the utility of the learnt prototypes for synthetic data generation.

\bibliography{refs}

\newpage
\appendix
\input{sections/appendix}

\end{document}

%% file: sections/introduction.tex
\begin{figure}[h]
    \centering
    \includegraphics[width=6cm]{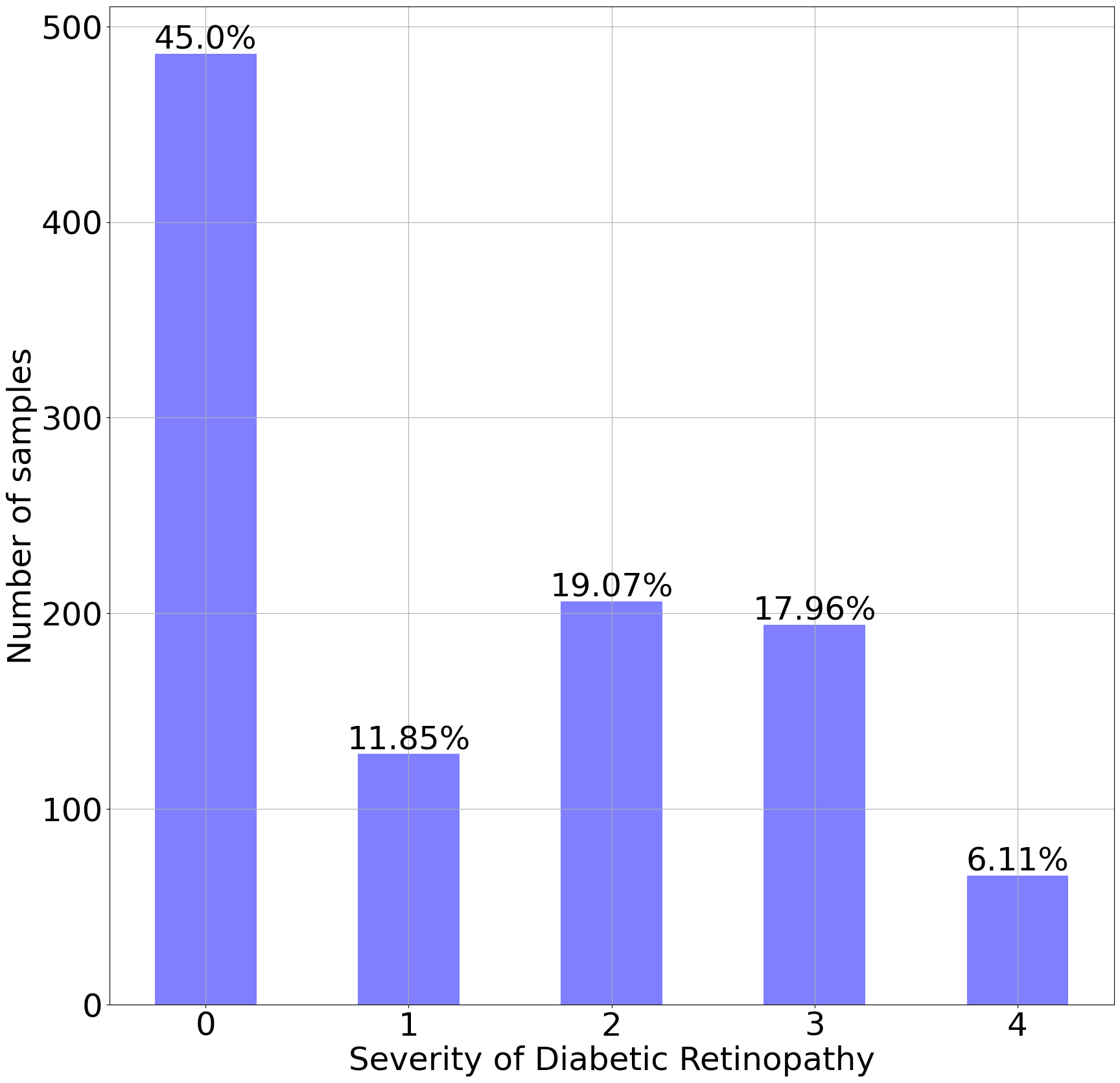}
    \caption{Class-wise data distribution from the RetinaMNIST dataset. Severity-1 Diabetic Retinopathy (DR) makes up 45\% of the dataset, whereas only $\sim$6\% (and only 66 images) of the data belongs to Severity-5 DR}
    \label{retina_class_wise}
\end{figure}

In recent years, the integration of deep learning techniques in medical and healthcare applications has exhibited remarkable progress, offering promising avenues for enhanced diagnostic and prognostic capabilities. However, the efficacy of these methods relies heavily on large annotated datasets. Accessing and labelling large quantities of medical images bears humongous costs in terms of the time and medical expertise required. \cite{li2021systematic} conducted a systematic study of over 300 medical imaging datasets reported between 2013 and 2020 and identified \textit{data scarcity} as the major bottleneck in the adoption of deep learning for medical image analysis. The authors advocate for the development of models that can operate effectively with less data. In addition to scarcity, medical imaging datasets are often characterized by severe class imbalance, with a single/few classes constituting majority of the dataset. For example, Fig.~\ref{retina_class_wise} shows the class-wise number (and percentage) of datapoints for RetinaMNIST, a dataset for grading the severity of diabetic retinopathy (DR) in the MedMNIST benchmark \cite{medmnistv1, medmnistv2}. Out of $\sim$1100 images, $\sim$500 images belong to Grade-1 DR, whereas only 66 images belong to Grade-5 DR. A neural network must be able to deal with such class-imbalance effectively in order to have a good performance at the task.

In addition to these challenges, for a neural network to be applicable in a medical setting, it is imperative that it can \textit{explain} its predictions. Explainability enhances trust in the model's predictions, which is crucial in a medical context. Current efforts in explainability include \textit{post hoc} methods, which explain the predictions of a trained black-box model by perturbing model-parameters or input/output pairs. Such methods include GradCAMs \cite{gradcam}, Guided BackProp \cite{guided_backprop}, Gradient SHAP \cite{shap} and LIME \cite{lime}, among other methods. \cite{explainability_eval} evaluate 16 such post-hoc explainability methods on whether they can meet clinical requirements on a multi-modal brain tumour grading task. They conclude that these methods fail to be \textit{faithful} to the model decision process at the feature-level and therefore, cannot be deployed directly in a medical setting. \cite{explainable-by-design} also point out the discrepancies among various post-hoc methods when applied to the same input and model, advocating for a shift toward explainable-by-design models for human-centric explainability.

In summary, medical image datasets are small and suffer from heavy class imbalance. Additionally, current post-hoc explainability methods are not reliable for clinical use. Hence, a neural network that is used for medical image classification must be - 
\begin{itemize}
    \item Data-Efficient (It must be able to learn from limited-data)
    \item Robust to class-imbalance
    \item Able to provide faithful and human-interpretable explanations
\end{itemize}

To tackle the aforementioned problems simultaneously, we introduce PRECISe, Prototype-Reservation for Explainable Classification under Imbalanced and Scarce-Data Settings. In summary, we make the following contributions -
\begin{itemize}
    \item We propose PRECISe, an explainable-by-design neural network that works well with limited and highly imbalanced training data
    \item We extensively evaluate the performance of our framework on various aspects - overall performance, data-efficiency and robustness to class-imbalance.
    \item We provide case-studies to demonstrate the explaibability of the proposed method and highlight the ease of human-interpretation of the provided explanations.
\end{itemize}

\subsection*{Generalizable Insights about Machine Learning in the Context of Healthcare}
We propose a framework, PRECISe, for training ML models which generalize well on unseen data with very limited ($<$60 images) training data. Our solution is robust to the underlying class imbalance present in medical image datasets, and provides human-interpretable explanations. Upon training on 50-60 images, PRECISe achieves $\sim$ 4.5\% gains over existing methods. Exaplainability is built into in the proposed framework, which aids in providing consistent and faithful explanations to the human user.


%% file: sections/relatedWork.tex
\textbf{Data Imbalance and Long-tailed distribution.}
Several approaches have been proposed to tackle the problem of imbalanced datasets in machine learning. \cite{ldam, tsc, gcl}. 
\cite{tsc} attempt to enhance the uniformity of imbalanced feature spaces by distributing features of different classes uniformly on a hypersphere. \cite{gcl} propose gaussian clouded logit adjustment via large amplitude perturbation, thereby making tail class samples more active in the embedding space. \cite{ldam} propose to replace the cross-entropy loss during training with a label-distribution aware distribution margin loss in an attempt to generalize to tail classes. However, the scale of data on which such models have been trained and evaluated is much larger than typical medical image datasets. Their ability to handle class imbalance on low-data settings needs further evaluation.

\noindent
\textbf{Explainable ML in Healthcare.}
As stated earlier, many of the existing \textit{post-hoc} explainability methods such as GradCams \cite{gradcam} have been evaluated for their clinical relevance \cite{clinical-eval-explainability-1, clinical-eval-explainability-2, clinical-eval-explainability-3}, and have been found to have limited relevance \cite{clinical-irrelevance}, with standard methods often highlighting the high-frequency regions in the image \cite{clinical-eval-explainability-1}. Consequently, alternative explainability methods have been proposed. \cite{visual-explainations} propose ensembling an adversarial classifier to generate visual explanations for diabetic retinopathy grading. \cite{prototype-dl} propose a prototype-based model which provides explanations through case based reasoning. However, their utility in scarce and imbalanced data settings has not been validated. 

Motivated by this, we propose PRECISe, an explainable-by-design model, which performs well under imbalanced and scarce-data settings.




%% file: sections/methodology.tex
\subsection{Network Architecture}
\begin{figure}
    \centering
    \includegraphics[height = 9cm, keepaspectratio]{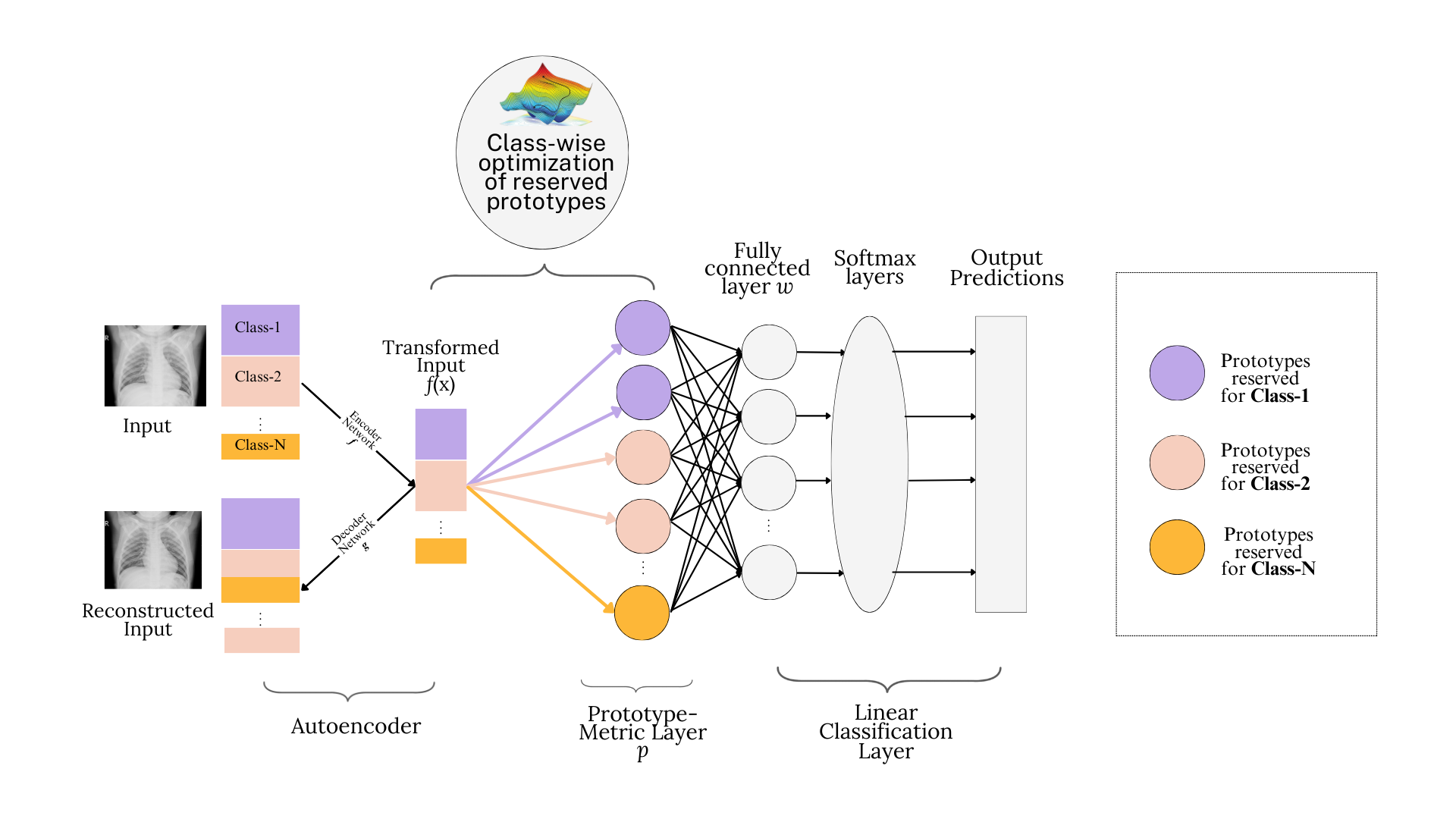}
    \caption{Overview of Prototype-Reservation}
    \label{methodology}
\end{figure}

The proposed model consists of 3 components - an auto-encoder, made up of an encoder $f$ and a decoder $g$, a prototype-metric layer $p$ and a linear classification layer $w$. The encoder transforms inputs into a compressed lower-dimensional latent space, and the decoder reconstructs the input from the corresponding space. The prototype-metric layer consists of two components - learnt \textit{prototypes} which resemble the representative samples in the training data and a metric layer, which transforms the encoding of an input (under the encoder) into a new $metric$ space. It must be noted that the prototypes reside in the latent space of the autoencoder ($R^d$). An input encoding ($\in R^d)$ is transformed into the $metric$ space by calculating the Euclidean distance of the encoding from each prototype, i.e. given an input encoding $x$ and $m$ prototypes, the prototype-metric layer transforms $x$ into $\hat{x}$ such that,
\begin{equation}
    \hat{x} = p(x) = [~||prototype_i - x||_2~] \forall i \in [1...m]
\end{equation}

The output from the prototype-metric layer is an $m$ dimensional vector, where $m$ is the number of prototypes in the layer. The linear classification layer generates a probability distribution over the $N$ output classes from this $m$-dimensional transformed vector in the $metric$ space. It consists of a fully-connected layer and a softmax layer. All the components of the network are trained jointly, as shown in Fig.~\ref{methodology}. Given that the prototypes reside in the latent space of the auto-encoder, their visualization becomes straightforward through reconstruction in the input space using the decoder. Explanatory features are inherent in this model, as the distance to each visualizable prototype for any input image can be readily observed. This facilitates a simple evaluation of the decision-making process to ascertain its validity. It is important to emphasize that the architecture of this network closely mirrors the one proposed by \cite{prototype-dl}. However, the novelty of our proposed method lies in the refinement of network optimization, specifically related to prototype synthesis, as elaborated below.

\subsection{Network Optimization - Prototype Reservation}
Since the network consists of 3 units, the loss function used to optimize the network is also made up of 3 components, the auto-encoder loss, the classification loss and the prototype loss. The auto-encoder loss (AE-Loss) is the (standard) Mean Squared Error Loss, meant to ensure faithful reconstructions of the input from its encodings. Let $D = {(x_i, y_i)}_{i=1}^{i=n}$, where $x_i$ is the input image and $y_i \in [1...N]$ be the training dataset and $\Tilde{x_i}$ is the reconstructed input, corresponding to input $x$, then the AE-Loss is expressed as -  

\begin{equation}
    loss_{AE} = \frac{1}{n} \Sigma_{i=1}^{i=n}||~\Tilde{x_i} - x_i~||_2^2; ~~\Tilde{x_i} = g ( f (x_i)
\end{equation}

In order to ensure optimal classification performance, the model is optimized with the standard Cross-Entropy loss, i.e. if $\hat{y_i}$ are the probability predictions corresponding to training datapoint $(x_i,y_i)$, the classification loss is expressed as -


Finally, the prototype loss is required to guarantee alignment between the prototypes and training data. This is done by ensuring that all training datapoints are $close$ to at least one prototype and that all prototypes are $close$ to at least one training datapoint. $Proximity$ is considered in the latent space. This idea was proposed by \cite{prototype-dl}. The first component of the prototype loss clusters similar datapoints close to the same prototype and the second component ensures that the prototype is a $summary$ of cluster around it. However, adopting this method for imbalanced data leads to all prototypes centered around the majority class, and no prototype corresponding to the minority class. We perform \textbf{prototype-reservation} in order to ensure that the network is robust to class-imbalance. Instead of optimizing the prototypes at a dataset-level, we propose optimizing the prototypes at a class-level, i.e. we minimize the distance between prototypes $reserved$ for that class and the datapoints corresponding to that class. This ensures that irrespective of the amount of training data, each class is adequately represented in the prototype-metric layer, and therefore robust to class-imbalance. If $PR_j$ represents a set of $d$ prototypes $reserved$ for $Class-j$, where $d$ is a hyperparameter, then the prototype-loss is expressed as - 
\begin{align*}
    L_{p} = 
    \frac{1}{n} 
    \Sigma_{i=1}^{i=n} 
    \Sigma_{j=1}^{j=N} 
    min_{proto_k \in PR_j}
    \mathbbm{1} [y_i==j] . || proto_k - x_i ||_2\\ 
    + \Sigma_{j=1}^{j=N} 
    \frac{1}{d*N} 
    \Sigma_{proto_k \in PR_j} 
    min_{i \in [1..n]} 
    \mathbbm{1} [y_i==j] . || proto_k - x_i ||_2
\end{align*}

The first term ensures that all datapoints labelled as $j$ are $close$ to at least one of the prototypes \textbf{\textit{reserved}} for class-$j$, whereas the second term ensures that all prototypes \textbf{\textit{reserved}} for class-$j$ are close to at least one of the datapoints labelled as $j$. Running this optimization for all classes present in the dataset ensures prototypes for each class are generated, i.e. $\forall j \in [1..n]$.

The complete loss function is expressed as, 
\begin{equation}
    loss = loss_{class} + \lambda_1 * loss_{AE} + \lambda_2 * l_p
\end{equation}
where $\lambda_1$ and $\lambda_2$ are coefficients used to control the contribution of the auto-encoder and prototype-generation in the model optimization, and are empirically identified. We use $\lambda_1 = 1$ and $\lambda_2 = 0.001$ for all our experiments. A good balance between these 3 terms leads to a network that has visualizable prototypes corresponding to all classes, and therefore, better classification accuracy.

%% file: sections/experiments.tex
\subsection{Datasets} 
\label{sec:datasets}
We evaluate PRECISe on 2 imbalanced medical imaging datasets - the Pneumonia \cite{chest-xray} and Breast Ultrasound Image (BUSI) \cite{busi} datasets.The pneumonia dataset consists of a total of 5232 Chest X-Ray images from patients (1349 normal, 3883 depicting pneumonia) for training and 624 chest X-rays for testing (234 normal and 390 depicting pneumonia). The dataset consists of AP view pediatric chest x-rays of patients (1-5 yrs) from the Guangzhou Women and Children’s Medical Center, Guangzhou. Low-quality / unreadable scans were subsequently removed. The ground truth for the images were obtained by grading by two expert physicians and verified by a third expert. Images are of variable size with the mean size being (1320, 968). The goal is to classify chest X-rays depicting pneumonia from normal images. The BUSI dataset consists of 780 images of breast ultrasound in women, with 487 scans with benign tumours, 210 with malignant tumours and 133 normal scans. The breast ultrasounds were collected from women between 25-75 years of age at the Baheya Hospital for Early Detection \& Treatment of Women's Cancer, Cairo, Egypt using the LOGIQ E9 ultrasound and LOGIQ E9 Agile ultrasound system. The average size of the images is 500x500 pixels. Images were annotated by radiologists in the same hospital. The small size of these datasets, along with the imbalance in different classes make them ideal for evaluation of PRECISe. While the Pneumonia dataset provides a separate test set, we randomly split 20\% of the BUSI dataset and use it as a dedicated test set.


\subsection{Baselines and Evaluation}

We compare the performance of PRECISe against 3 baselines - a standard ResNet-50 \cite{resnet50} model tuned in a fully supervised manner, the original prototypes method proposed by \cite{prototype-dl} and LDAM \cite{ldam}, which uses a Label-Distribution Aware Margin Loss to generalize to an imbalanced long-tailed distribution in the training set.
As mentioned earlier, we evaluate PRECISe on 3 aspects - data-efficiency, generalization to minority classes and explainability. To estimate the data-efficiency of the proposed method, we take two measures - 
\begin{itemize}
    \item We choose datasets with limited number of images available for training. The scale of these datasets is a fraction of the size of datasets currently used to train neural networks for image recognition such as the ImageNet \cite{imagenet} or LAION-400M \cite{laion} datasets
    \item We additionally report performance by training models on subsets of varying sizes (1, 5, 10, 25, 50, 100 \% of the entire set). The 1\% split is omitted for the BUSI dataset due to its smaller size.
\end{itemize}

In addition to accuracy, we also report the F1-score averaged over all classes as a performance measure.


To evaluate the generalization of the proposed method to minority classes, we also report the classwise accuracies of all methods. Classwise accuracy is calculated as the proportion of datapoints in each class that the model classifies correctly. We report these for the smallest subset of both datasets (1\% subset for Pneumonia, 5\% for BUSI) to measure the ability of the model to generalize on minority classes, even with very less data. All results are reported as the mean of 3 independent runs. Different subsets are chosen for the 3 runs when training on $<$100\% of the dataset, however, the subsets are consistent across all methods. 
Finally, we also present an explainability case-study to demonstrate the utility of visual explanations obtained from the model. 


\subsection{Implementation}
\label{implementation}
We adopt the ResNet-50 backbone \cite{resnet50} for all methods, specifically, the encoder for PRECISe and Prototypes \cite{prototype-dl} is a ResNet-50 model, with the output projected to a 256-dimensional embedding using a linear layer. The decoder for the proposed method is a fully convolutional network, consisting of upsampling and convolutional operations. Images for both datasets are resized to 224x224 to be used with ResNet50. We do not apply data augmentations for training. We only normalize with the ImageNet mean and standard deviation before processing the image. The number of prototypes reserved for each class was chosen based on experimentation with different number of prototypes reserved for each class (10\% data) (details in Appendix). We reserve 2 prototypes for each class in the Pneumonia dataset, and 3 prototypes for each class in the BUSI dataset. Additionally, we use a weighted cross-entropy loss function for the classifier - this helps in preventing the decoder from learning to decode only the majority class. The weights of individual classes are inversely proportional to their frequency of occurrence in the dataset. The model is optimized with the Adam optimizer with a constant learning rate of 1e-3 and a weight decay coefficient of 1e-4. \footnote{The code implementation for the proposed method is publicly available \href{https://github.com/ganatra-v/PRECISe}{here} }

We empirically found that initializing the architecture with pretrained weights on the ImageNet dataset \cite{imagenet} yields better accuracies. This has also been observed in prior works \cite{respirenet}. Possibly, the network's understanding of shape and preliminary structures help in generalization to medical images as well. Hence, all ResNet-50 instances are initialized with pretrained ImageNet weights. Consequently, we call the supervised ResNet-50 baseline as FT (Finetuning) as it is pretrained on the ImageNet dataset and is being finetuned on the current datasets.

%% file: sections/results.tex
\subsection{Results and Discussion}

\begin{table}
\centering
\begin{tabular}{|c|c|c|}
\hline
Method         & Accuracy               & Mean F1 score          \\ \hline
FT             & 89.476 $\pm$ 0.421         & 88.096 $\pm$ 0.522          \\ \hline
LDAM           & 87.393 $\pm$ 0.659         & 85.604 $\pm$ 0.907          \\ \hline
Prototypes     & 91.293 $\pm$ 0.529    & 90.726 $\pm$ 0.526          \\ \hline
PRECISe (ours) & \textbf{92.041 $\pm$ 0.151} & \textbf{91.340 $\pm$ 0.053} \\ \hline
\end{tabular}
\caption{Overall accuracy and Mean F1-scores on the Pneumonia dataset. PRECISe (ours) outperforms all baselines.}
\label{pneumonia-perf}
\end{table}

\begin{table}
\centering
\begin{tabular}{|c|c|c|}
\hline
Method         & Accuracy               & Mean F1 score          \\ \hline
FT             & 69.427 $\pm$ 0.520          & 54.143 $\pm$ 1.756          \\ \hline
LDAM           & 80.255 $\pm$ 0.520          & 32.718 $\pm$ 7.054          \\ \hline
Prototypes     & 87.898 $\pm$ 0.520          & 86.580 $\pm$ 0.863          \\ \hline
PRECISe (ours) & \textbf{88.747 $\pm$ 0.601} & \textbf{86.939 $\pm$ 1.482} \\ \hline
\end{tabular}
\caption{Overall accuracy and Mean F1-scores on the BUSI dataset. PRECISe (ours) outperforms all baselines.}
\label{busi-perf}
\end{table}

\textbf{Overall Performance} : Tables~\ref{pneumonia-perf} and \ref{busi-perf} show the performance of the 
proposed method and baselines on the Pneumonia and BUSI datasets respectively. The proposed method, PRECISe, outperforms all baselines, achieving the highest accuracy of 92.04\% and 88.75\% on the Pneumonia and BUSI datasets, respectively.

\begin{figure}
    \centering
    \includegraphics[width = 9cm]{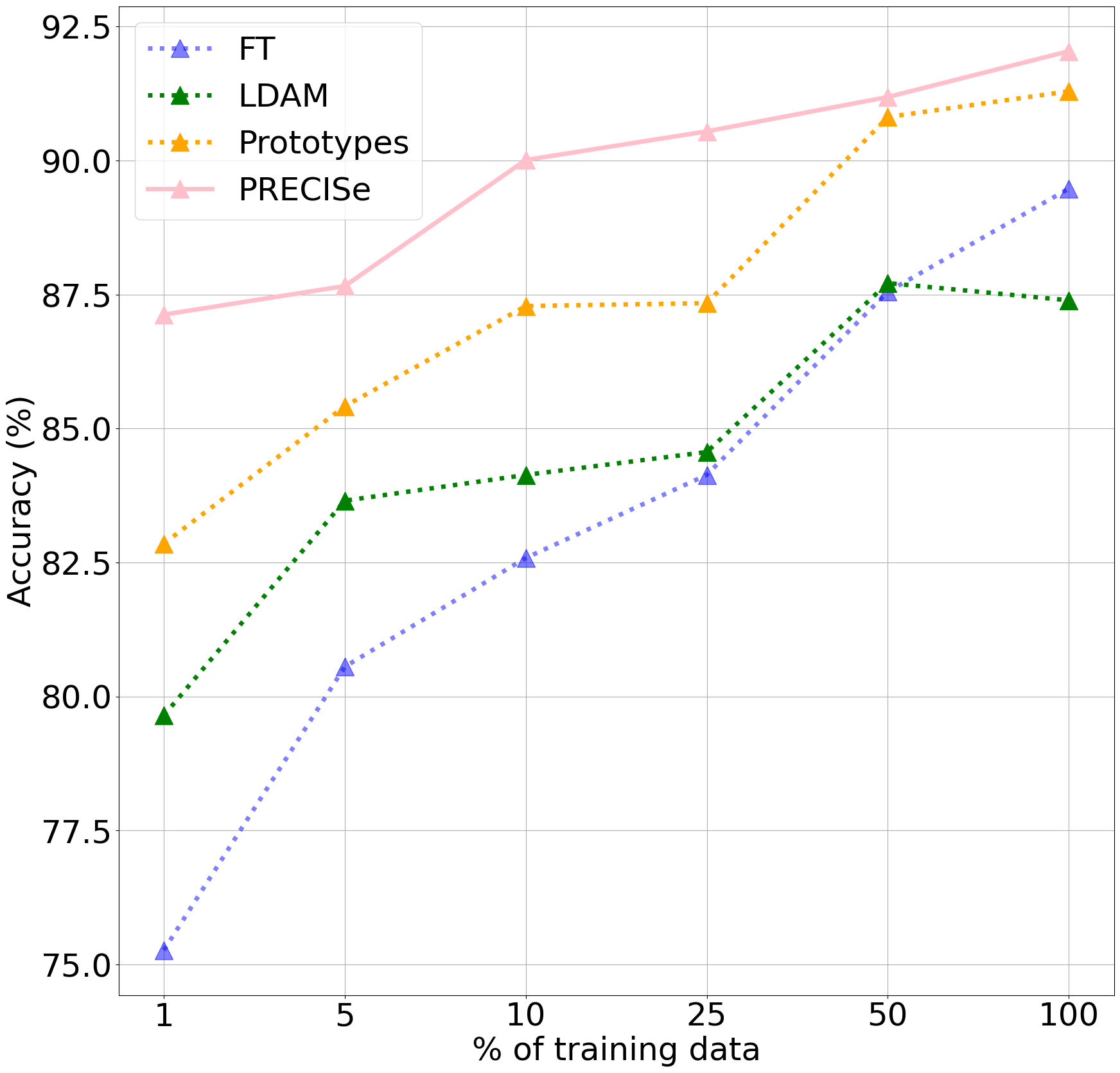}       
    \caption{Performance on subsets of varying sizes of the Pneumonia dataset. PRECISe (ours) shows excellent performance retention with reducing training set sizes}
    \label{pneumonia-scarcity}     
\end{figure}

\begin{figure}
    \centering
    \includegraphics[width = 9cm]{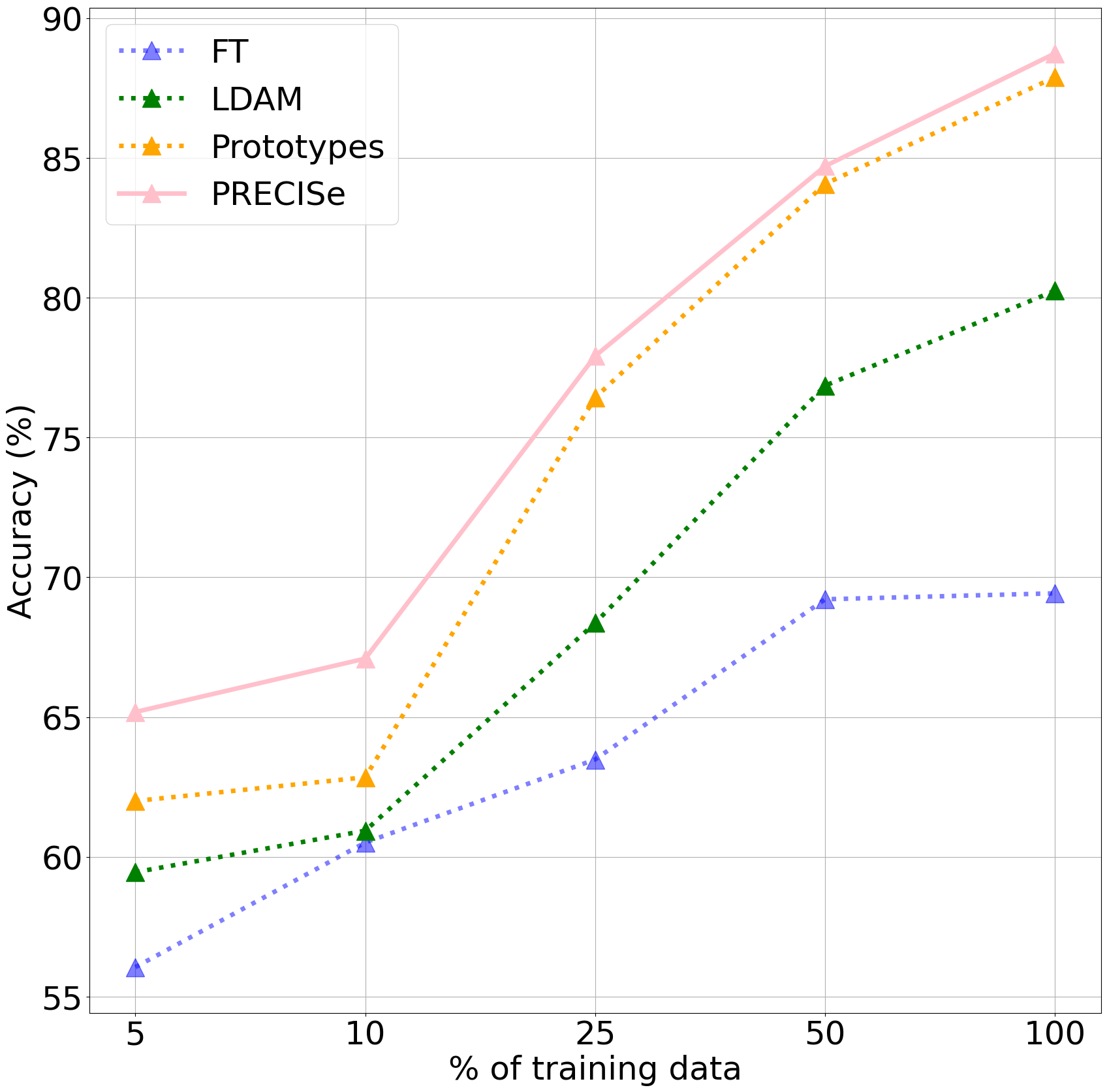}       
    \caption{Performance on subsets of varying sizes of the BUSI dataset. PRECISe (ours) shows the best performance at all training set sizes.}
    \label{busi-scarcity}        
\end{figure}

\noindent
\textbf{Data Efficiency.} Figures~\ref{pneumonia-scarcity} and \ref{busi-scarcity} show the performance of all methods on subsets of varying sizes of the Pneumonia and BUSI datasets, respectively. Again, PRECISe outperforms all methods and achieves highest accuracies. Additionally, whereas other methods suffer from a significant performance drop upon reducing training set sizes, PRECISe shows excellent performance retention. For example, using only 1\% of training images ($<$60) in the Pneumonia dataset, PRECISe achieves an accuracy of 87.13\%, as compared to a 92.04\% accuracy using the entire Pneumonia dataset. The performance drop is much smaller than that seen in FT (89.48\% acc. @ 100\% data vs 75.28\% acc. @ 1\% data) or LDAM (87.39\% acc. @ 100\% data vs 79.647\% acc. @ 1\% data). We hope that a model which is able to achieve $\sim$87\% accuracy using $<$60 labels might go a long way in reducing the time/effort spent by doctors and medical professionals in annotating medical data.

It must be noted that both Prototypes \cite{prototype-dl} and PRECISe (ours) significantly outperform other methods. We speculate that this is due to the $prototype-metric$ layer. In the typical paradigm of training neural networks, a model is expected to learn representations which compress the information in the input data as well as separate representations of difference classes well. We speculate that the transformation of an input encoding into the $metric$ space by finding the Euclidean distance from the prototypes aids the separation of the representations, hence, it is easier for the model to summarize the information about the training distribution in the form of learnt prototypes.

\begin{figure}
    \centering
    \includegraphics[height = 6.5cm]{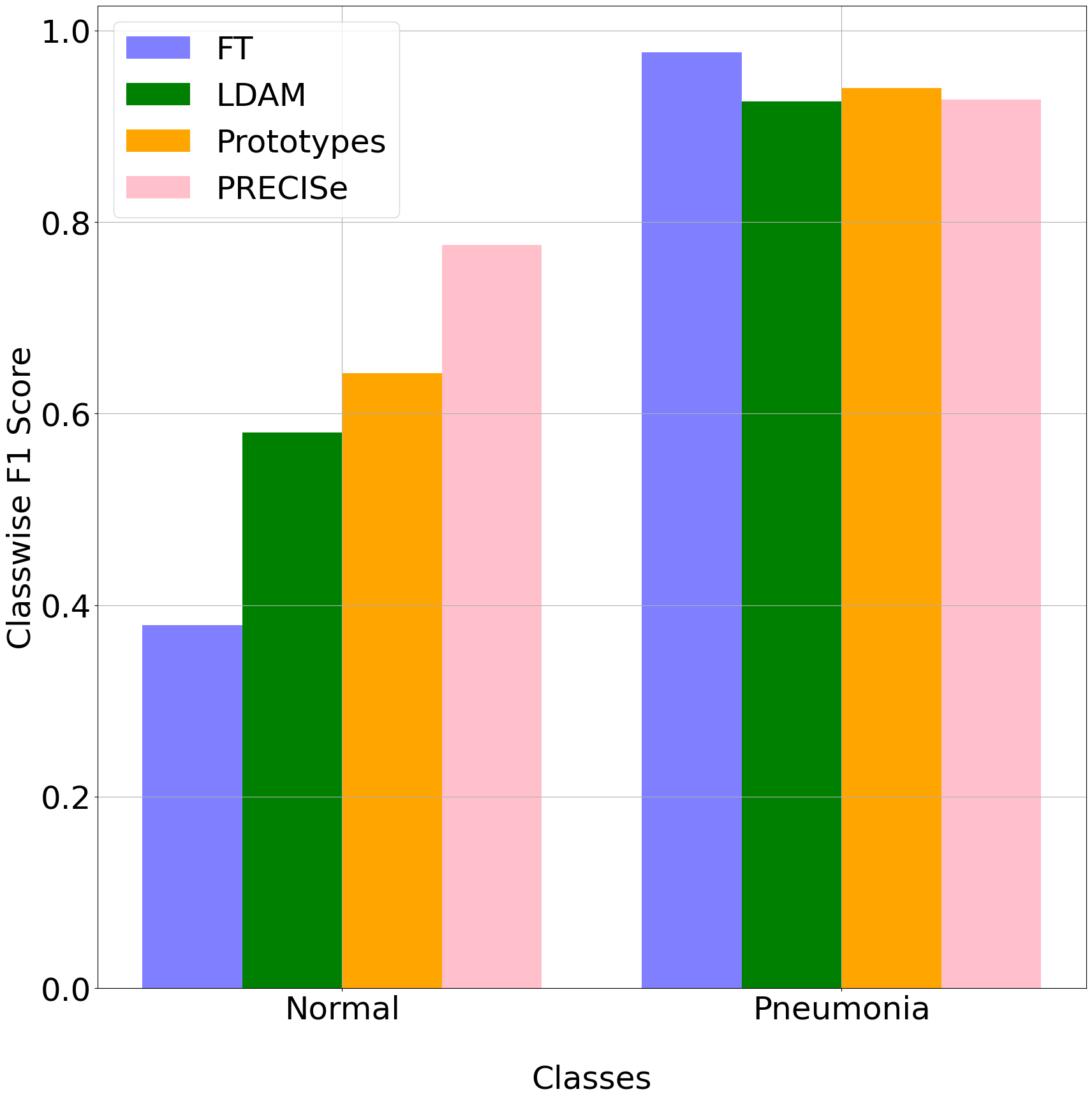}
    \caption{Classwise performance on the Pneumonia dataset. PRECISe (ours) displays superior performance on the minority classes.}
    \label{pneumonia-classwise}
\end{figure}

\begin{figure}[h]
    \centering
    \includegraphics[height = 8cm]{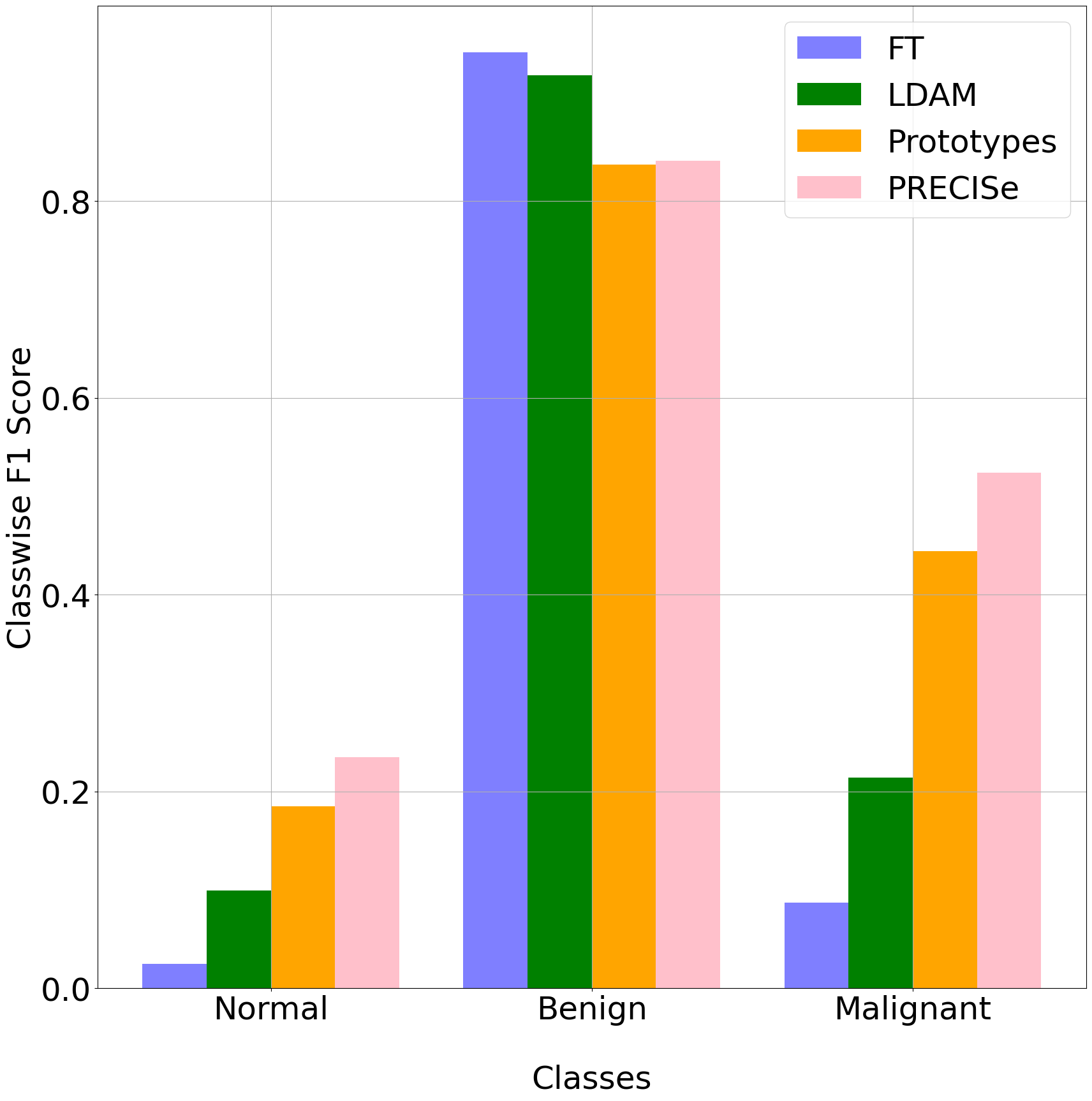}
    \caption{Classwise performance on the BUSI dataset. PRECISe (ours) displays superior performance on the minority classes.}
    \label{busi-classwise}
\end{figure}

\textbf{Generalization to Minority Classes} - Figures~\ref{pneumonia-classwise} and \ref{busi-classwise} depict the classwise accuracy of each method on the Pneumonia and BUSI datasets, respectively. 

As summarized in Sec~\ref{sec:datasets}, the ``Normal'' class is the minority in the Pneumonia dataset, with the majority class having $\sim$3x data. Similarly, the ``Benign'' class is the majority class in the BUSI dataset with $\sim$3x more data than the ``Normal'' class. In addition to obtaining the best overall accuracies, PRECISe also achieves the best performance on the minority classes, with a comparable performance on the majority class. It must be noted that the results reported for Figures~\ref{pneumonia-classwise} and \ref{busi-classwise} are for the smallest subset, i.e. 1\% subset for the pneumonia set ($<$60 images) and 5\% split for the BUSI dataset ($<$35 images), while maintaining the class-imbalance ratio. Despite the very small sample size, PRECISe is able to identify 77.6\% of all normal chest X-rays of the Pneumonia dataset, as compared to a 37.9\% for FT and 58\% for LDAM. Similarly, PRECISe is able to identify 52.4\% of all Malignant tumours correctly in the BUSI dataset, as opposed to an 8.7\% by FT, 21.4\% by LDAM and 44.4\% by Prototypes. This highlights that fact that improved overall  performance of the proposed method is because of improved performance on the minority classes, as is desired.

\subsection{Explainability Case Study}
\begin{figure}
    \centering
    \includegraphics[width=15cm]{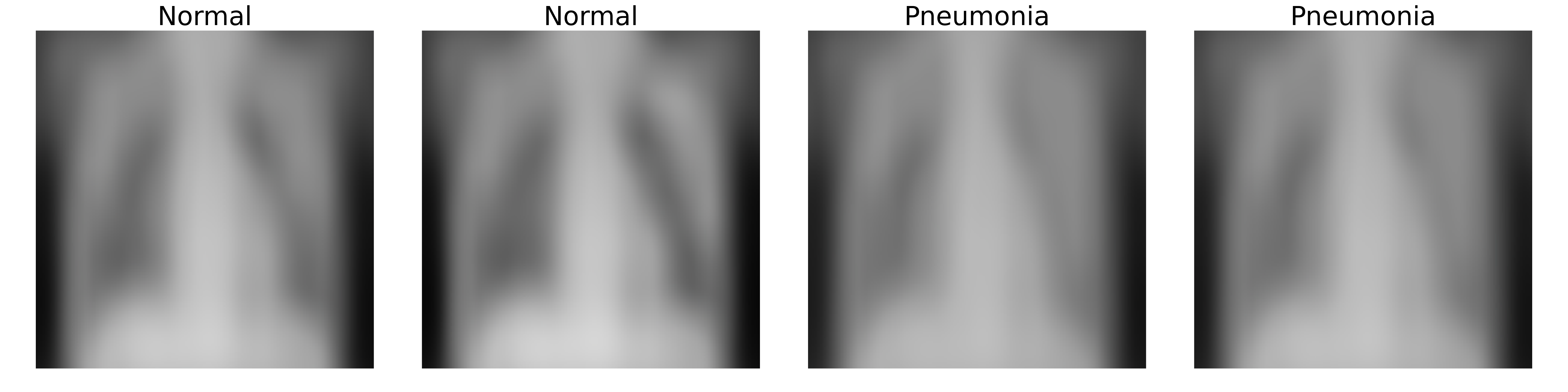}
    \caption{Prototypes generated during the training of PRECISe on the Pneumonia Dataset}
    \label{explainability-case-study-prototypes}
\end{figure}

\begin{table}
    \centering
    \begin{tabular}{|c|c|c|c|c|}
    \hline
&&&&\\    
    Prototypes $\rightarrow$
    &\includegraphics[width = 3cm, height = 3cm]{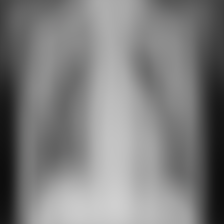}
    &\includegraphics[width = 3cm, height = 3cm]{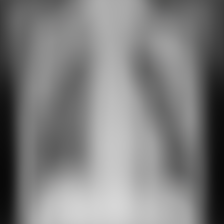}
    & \includegraphics[width = 3cm, height = 3cm]{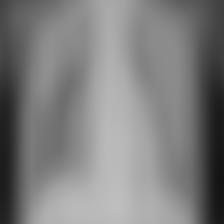}
    & \includegraphics[width = 3cm, height = 3cm]{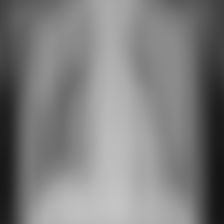}\\
    Sample Images $\downarrow$
    &(Normal)&(Normal)&(Pneumonia)&(Pneumonia)\\
    \hline
&&&&\\
    \includegraphics[width=3cm]{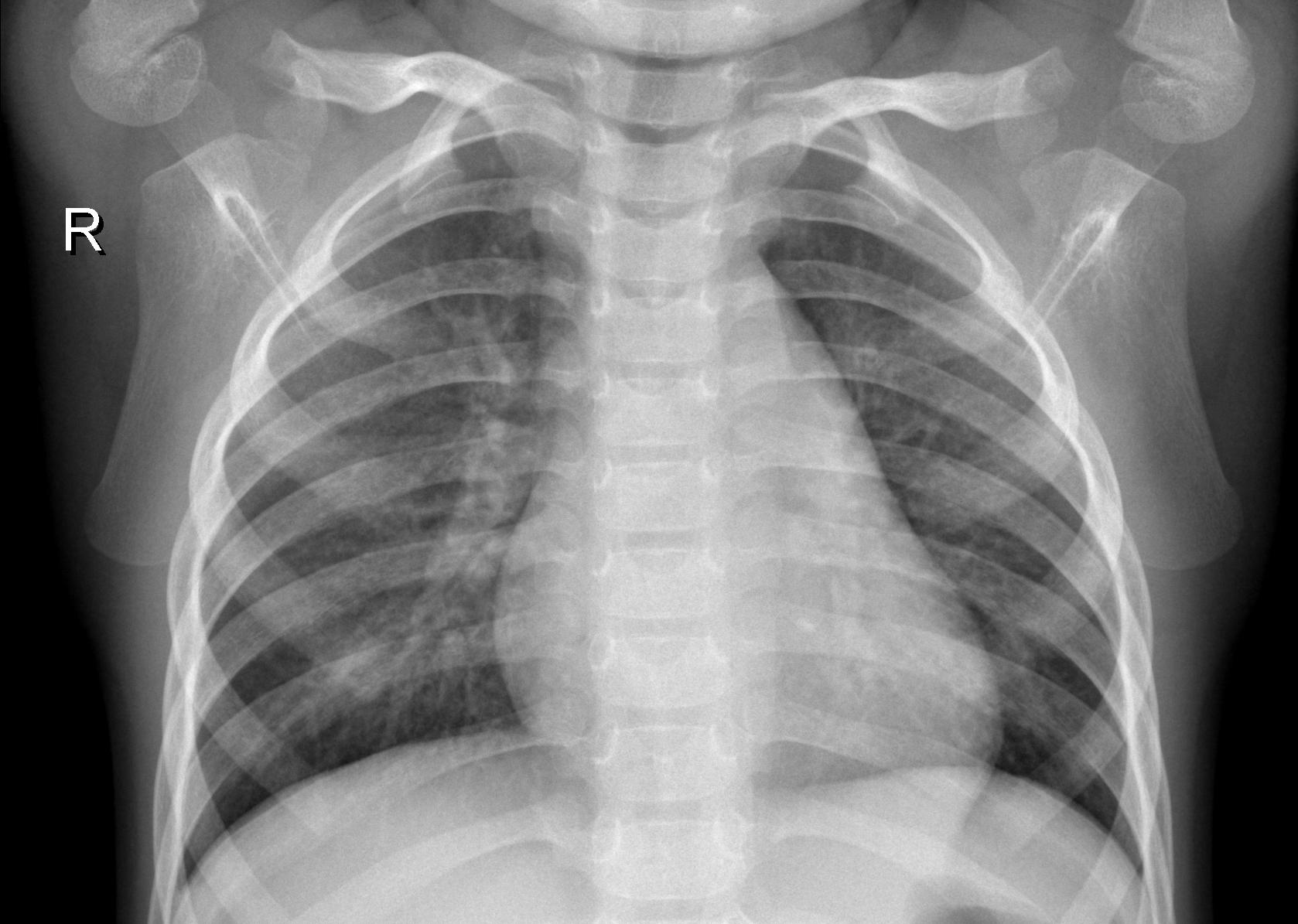} & \textbf{2.283} & \textbf{0.519} & 7.920 & 5.424\\
    (Normal) &&&&\\
    \hline
&&&&\\
    \includegraphics[width=3cm]{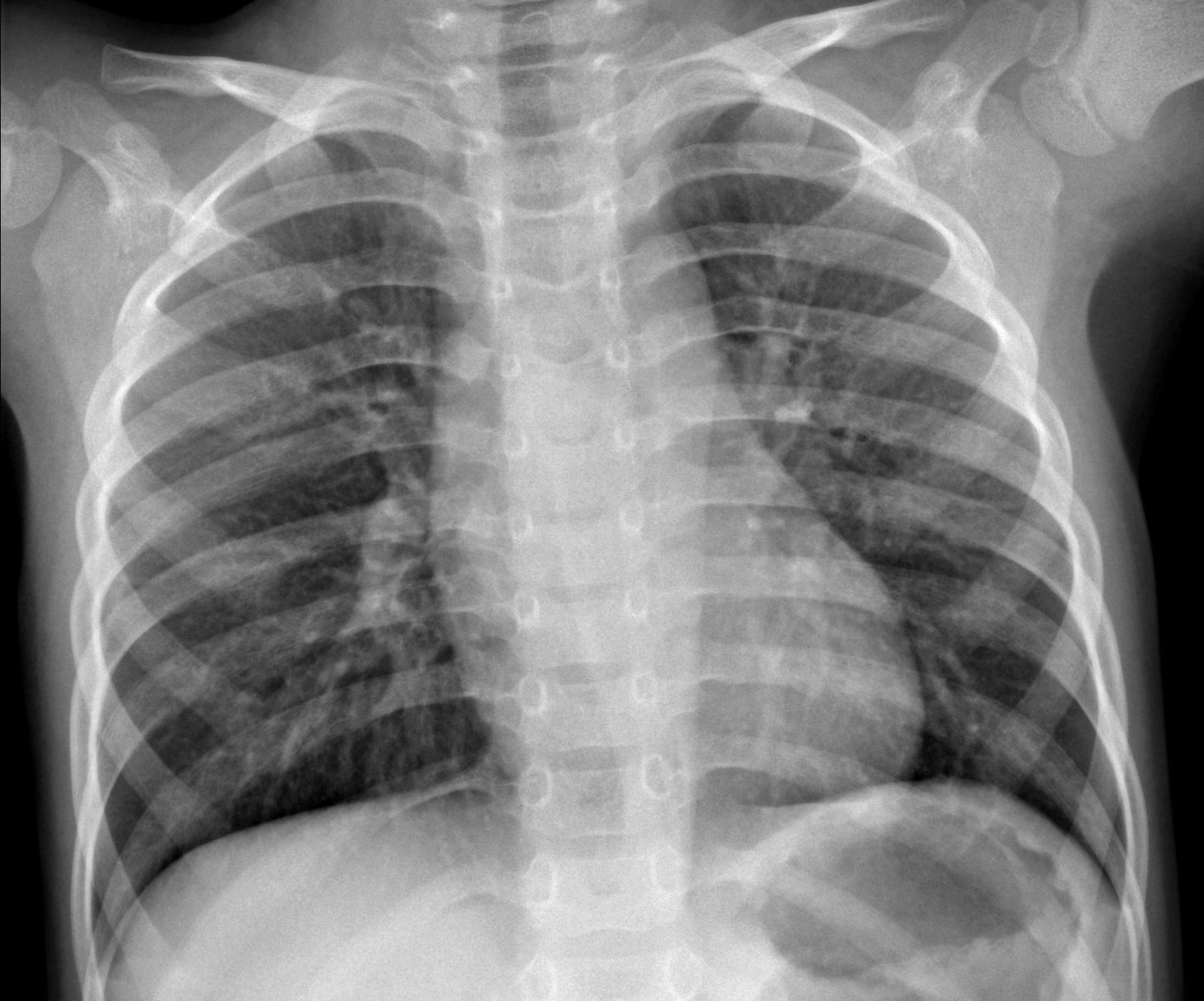} & \textbf{2.363} & \textbf{0.712} & 7.998 & 5.499\\
    (Normal) &&&&\\
    \hline
&&&&\\
    \includegraphics[width=3cm]{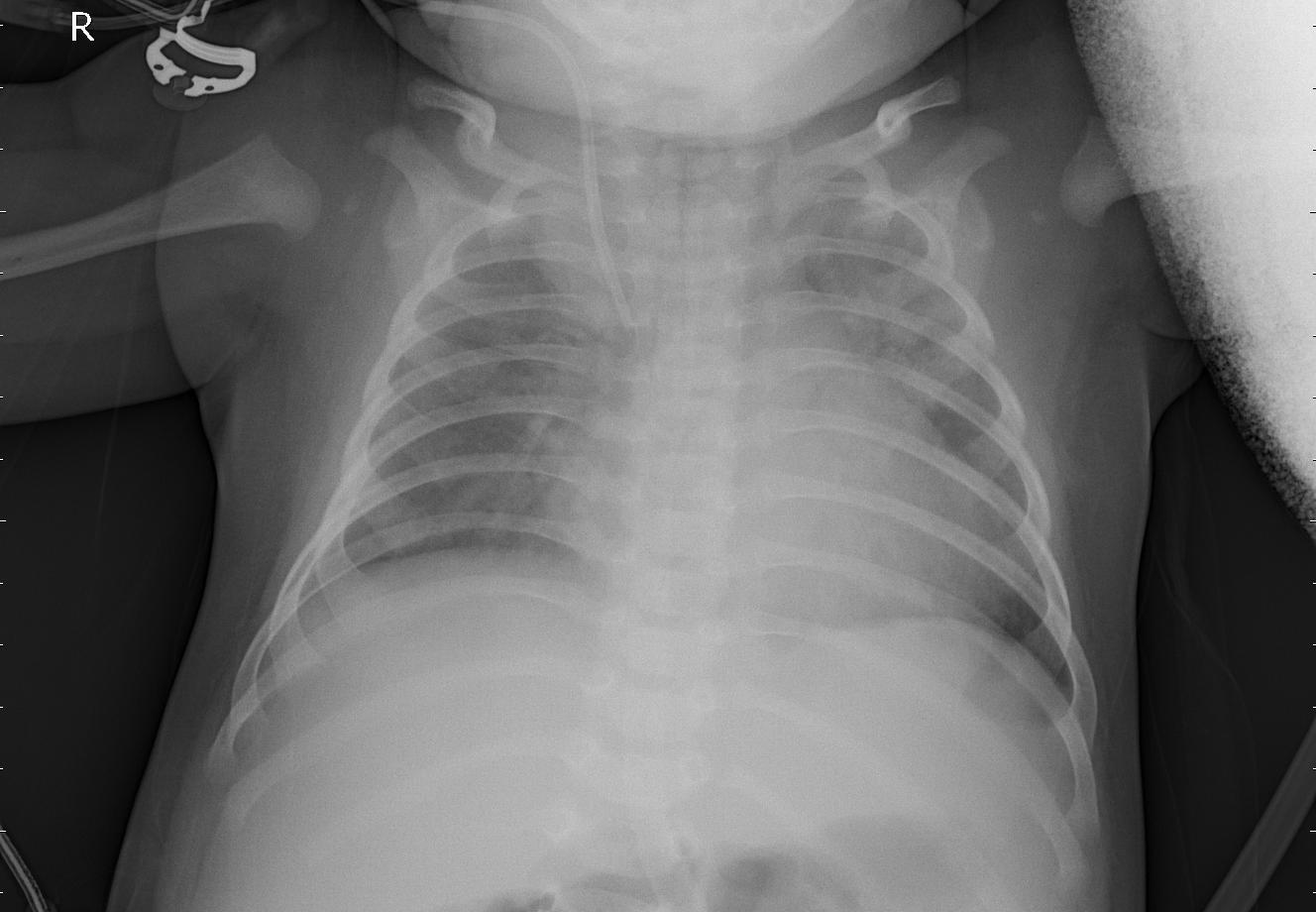}& 7.064 & 9.501 & \textbf{1.369} & \textbf{3.864}\\
    (Pneumonia) &&&&\\
    \hline
&&&&\\
    \includegraphics[width=3cm]{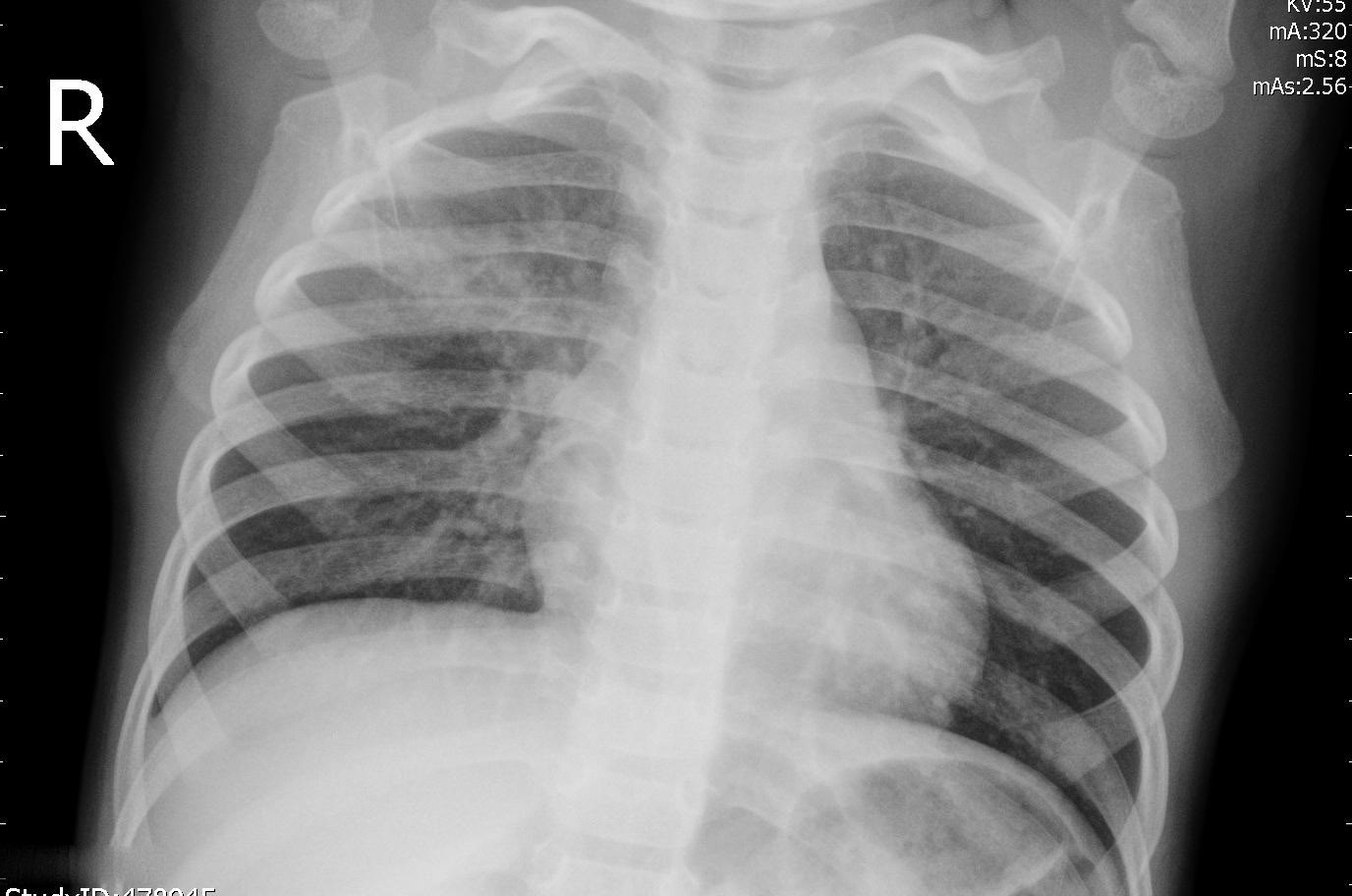} & 6.543 & 8.962 & \textbf{1.118} & \textbf{3.407}\\
    (Pneumonia) &&&&\\
    \hline
    \end{tabular}
    \caption{Euclidean distances of images (left) from prototypes (top) - A case-by-case examination of network explainability. Normal images are closer to normal prototypes, whereas images with pneumonia are closer to prototypes for the pneumonia class. For each image, the corresponding row is the input to the final linear layer, thereby using interpretable features for classification.}
    \label{case-by-case-explainability}
\end{table}

\begin{table}[h]
\centering
\begin{tabular}{|l|l|l|}
\hline
                                                           & \begin{tabular}[c]{@{}l@{}}Normal\\ Prototypes\end{tabular} & \begin{tabular}[c]{@{}l@{}}Pneumonia\\ Prototypes\end{tabular} \\ \hline
\begin{tabular}[c]{@{}l@{}}Normal\\ Images\end{tabular}    & \textbf{2.536}                                                       & 5.171                                                          \\ \hline
\begin{tabular}[c]{@{}l@{}}Pneumonia\\ Images\end{tabular} & 7.932                                                       & \textbf{3.499}                                                          \\ \hline
\end{tabular}
\caption{Average distance of all testing data from prototypes reserved for various classes. Data belonging to a particular class is closer to prototypes reserved for that class.}
\label{average-distance}
\end{table}

Fig.~\ref{explainability-case-study-prototypes} shows the prototypes that were learnt when training PRECISe on Pneumonia dataset. Prototypes reserved for normal class appear darker and those of pneumonia appear much paler. This is consistent with the fact that fluid in the lungs manifests as a pale white overlay on the lung region in the chest X-rays. Table.~\ref{case-by-case-explainability} shows the Euclidean distances of randomly chosen normal and pneumonia images (from the test set) on a case-by-case basis from the prototypes. It is distinctly seen that images of either class are much closer to the prototypes of the same class, than those of the other class. Additionally, for each image, the corresponding row is the input to the final linear classification layer. Fully interpretable features being used for classification enables us to examine (and interpret) the basis on the which the model has provided a decision. Moreover, due to the architectural design, the model is optimized to classify based on the ``right reasons". This fact is demonstrated in Table.~\ref{average-distance}, which shows the mean Euclidean distance of all the images in the test set to the prototypes of both classes. Once again, we observe that the datapoints from either class are closer to the prototypes of the corresponding class as compared to the alternative class, which proves the faithfulness of the human interpretable explainations provided by the model.

\subsection{Limitations}
We acknowledge a few key limitations of the proposed approach. Firstly, the nature of explanations provided by the model are based on whole-image similarity. Unlike existing methods such as GradCams \cite{gradcam} which highlight the activation regions in the image, PRECISe provides explanations in terms of visually similar images. Hence, interpreting the explanations may still require domain knowledge.
Secondly, the proposed setup has not been evaluated by medical professionals. End-to-end evaluation of the approach via user studies may help determine the clinical usability of the system.

%% file: sections/discussions.tex

%% file: sections/appendix.tex
\section*{Appendix}
\label{appendix}

\section*{Determining the number of prototypes reserved for each class}

As mentioned in Sec \ref{implementation}, the number of prototypes reserved for each class was chosen based on experimentation. We tried reserving 1-5 prototypes per class for both the Pneumonia and BUSI datasets using 10\% of the training data for experimentation. We found that reserving 2 prototypes per class for the Pneumonia dataset and 3 prototypes per class for the BUSI dataset to yield the best performance. Table \ref{num_prototype_selection} shows the performance of PRECISe upon reserving varying number of prototypes. Interestingly, we find that varying the number of prototypes reserved per class does not cause a significant difference in the performance on the Pneumonia dataset, but leads to considerable difference in performance on the BUSI dataset.

\begin{table}[h]
\centering
\begin{tabular}{|c|c|c|}
\hline
\#-prototypes & Pneumonia      & BUSI           \\ \hline
1             & 88.30          & 60.51          \\ \hline
2             & \textbf{90.01} & 62.42          \\ \hline
3             & 89.90          & \textbf{67.09} \\ \hline
4             & 89.26          & 61.14          \\ \hline
5             & 89.58          & 62.42          \\ \hline
\end{tabular}
\caption{Performance of PRECISe (ours) upon reserving different number of prototypes per class.}
\label{num_prototype_selection}
\end{table}